\documentclass[sigconf]{acmart}
\usepackage{balance}
\usepackage{pifont}
\usepackage{multirow}
\newcommand{\cmark}{\ding{51}}
\newcommand{\xmark}{\ding{55}}

\AtBeginDocument{%
  }

\copyrightyear{2025}
\acmYear{2025}
\setcopyright{acmlicensed}
\acmConference[MM '25] {Proceedings of the 33rd ACM International Conference on Multimedia}{October 27--31, 2025}{Dublin, Ireland.}
\acmBooktitle{Proceedings of the 33rd ACM International Conference on Multimedia (MM '25), October 27--31, 2025, Dublin, Ireland}
\acmISBN{979-8-4007-2035-2/2025/10}
\acmDOI{10.1145//3746027.3758221}

\usepackage{caption}
\begin{document}
	

\title{PhysLab: A Benchmark Dataset for Multi-Granularity Visual Parsing of Physics Experiments}


\author{Minghao Zou}
\affiliation{%
	\institution{Shandong University of Science and Technology}
	\city{Qingdao}
	\country{China}
}


\affiliation{%
	\institution{Cardiff University}
	\city{Cardiff}
	\country{UK}
}
\email{mhzou@sdust.edu.cn}

\author{Qingtian Zeng}
\affiliation{%
	\institution{Shandong University of Science and Technology}
	\city{Qingdao}
	\country{China}
}
\email{qtzeng@163.com}
\authornote{Corresponding authors.}

\author{Yongping Miao}
\affiliation{%
	\institution{Shandong University of Science and Technology}
	\city{Qingdao}
	\country{China}
}
\email{miaoyongping@sdust.edu.cn}

\author{Shangkun Liu}
\affiliation{%
	\institution{Shandong University of \\ Science and Technology}
	\city{Qingdao}
	\country{China}
}
\email{liushangkun@sdust.edu.cn}

\author{Zilong Wang}
\affiliation{%
	\institution{Shandong University of \\ Science and Technology}
	\city{Qingdao}
	\country{China}
}
\email{zlwang@sdust.edu.cn}

\author{Hantao Liu}
\affiliation{%
	\institution{Cardiff University}
	\city{Cardiff}
	\country{UK}
}
\email{liuh35@cardiff.ac.uk}

\author{Wei Zhou}
\affiliation{%
	\institution{Cardiff University}
	\city{Cardiff}
	\country{UK}
}
\email{zhouw26@cardiff.ac.uk}
\authornotemark[1]

\renewcommand{\shortauthors}{Minghao et al.}

\begin{abstract}
Visual parsing of images and videos is critical for a wide range of real-world applications. However, progress in this field is constrained by limitations of existing datasets: (1) limited annotation diversity, which limits the support for diverse vision tasks within a unified dataset; (2) insufficient coverage of domains, particularly a lack of datasets tailored for educational scenarios; and (3) a lack of explicit procedural guidance, with weak logical rules and insufficient representation of a structured task process. To address these gaps, we introduce PhysLab, the first dataset that captures students conducting complex physics experiments. The dataset includes four representative experiments that feature diverse scientific instruments and rich human-object interaction (HOI) patterns. PhysLab comprises 620 long-form videos and provides multi-granularity annotations that support a variety of vision tasks, including action recognition, object detection, HOI analysis, etc. We establish baselines and perform extensive evaluations to highlight key challenges in the parsing of procedural educational videos. We expect PhysLab to serve as a valuable resource for advancing comprehensive visual parsing, facilitating intelligent classroom systems, and fostering closer integration among computer vision, multimedia, and educational technologies. The dataset and the evaluation toolkit are publicly available at \href{https://github.com/ZMH-SDUST/PhysLab}{https://github.com/ZMH-SDUST/PhysLab}.
\end{abstract}




\maketitle

\section{Introduction}

Visual parsing is a fundamental research area in multimedia, encompassing a wide range of tasks such as action recognition, object detection, and HOI detection \cite{yu2022mm}. It plays a pivotal role in real-world applications, including surveillance, autonomous driving, and industrial inspection. In recent years, the field has achieved remarkable progress, driven by advances in deep learning and the availability of large-scale annotated datasets \cite{zhou2025perceptual}. However, this data-centric paradigm also exposes a strong reliance on high-quality and diverse annotations. Although existing datasets are large in scale, they remain short in supporting multi-task collaborative modeling and rich semantic understanding \cite{li2024finerehab}. Specifically, they are often tailored to single-task learning or provide only coarse-grained labels, lacking the structured and detailed annotations needed to jointly model actions, objects, and interactions \cite{kim2025visual}. Furthermore, most datasets are predominantly built around common scenarios, such as households \cite{tang2019coin}, streets \cite{liu2024trafficmot}, and kitchens \cite{lee2024error}, while neglecting cognitively demanding domains like education. As a result, many baseline models tend to prioritize static appearance cues or short-term visual patterns, falling short in modeling more explicit causal relationships, long-term temporal dependencies, and strict physical constraints that are characteristic of complex procedural tasks \cite{wu2024sportshhi}. These limitations hinder the development of models capable of deeper and human-like understanding.\par

\begin{figure*}
	\centering
	\includegraphics[width=0.83\textwidth]{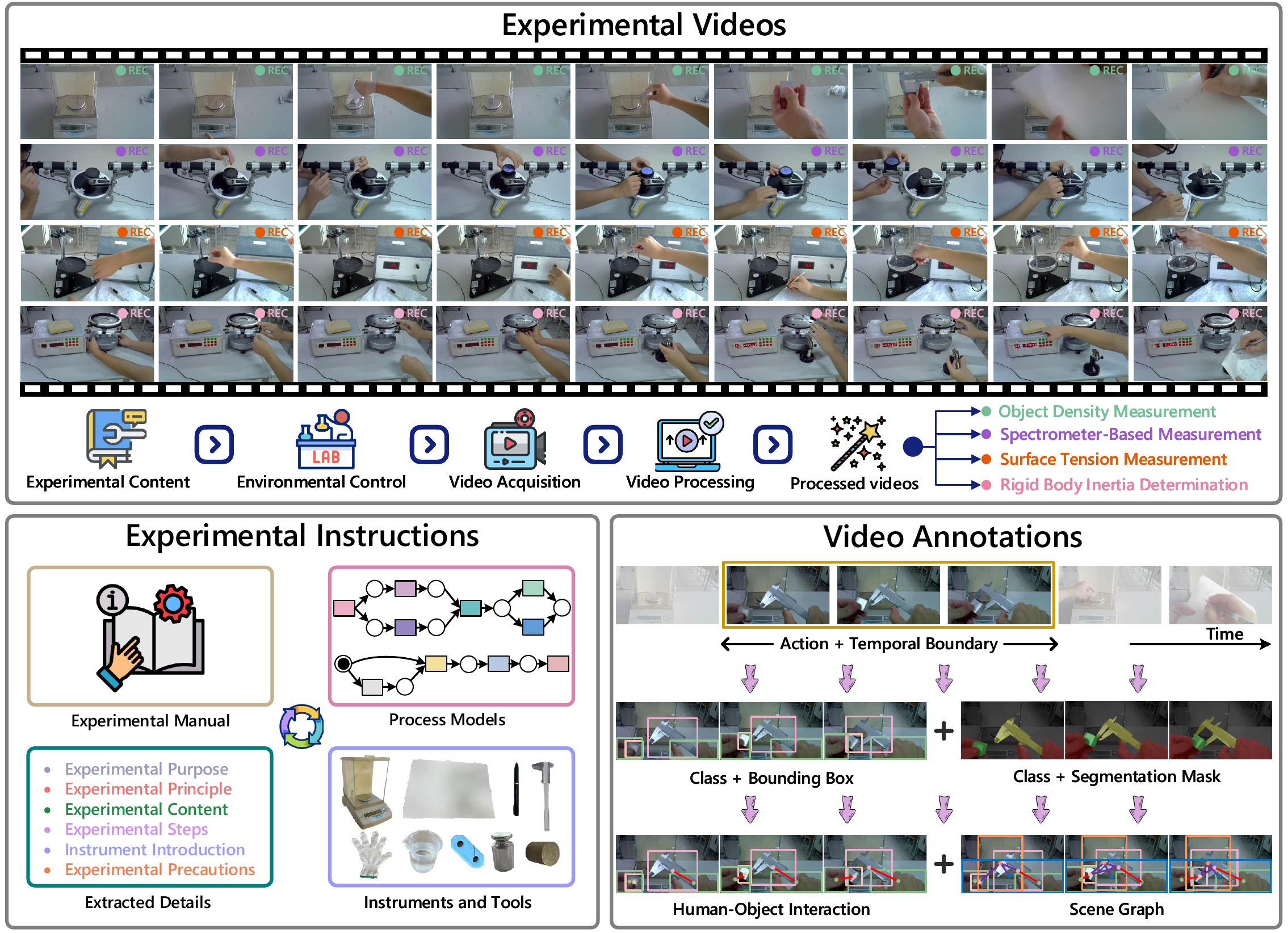}
	\caption{An overview of the content of our proposed PhysLab dataset. PhysLab focuses on experimental tasks conducted in complex physics laboratory settings. 1) The top row shows the data collection process and representative data samples. 2) The second row shows the distinctive operational instructions associated with physics experiments, including detailed procedures, instrument usage, and process models in Petri Net format (lower-left), along with multi-granularity annotations (lower-right).}
	\label{fig:intro}
\end{figure*}

Compared to conventional visual scenes, physics experiments exhibit significantly greater spatiotemporal complexity and cognitive demands \cite{latif2024physicsassistant}. Experimental procedures typically involve multi-step operations, instrument configurations, and parameter adjustments, all governed by strict task dependencies and procedural order \cite{zou2024weakly}. These tasks also feature frequent and detailed human-instrument interactions, including tool switching and feedback observation. Such interactions require nuanced recognition and cross-modal reasoning \cite{huang2024egoexolearn}. Additionally, experimental processes are often underpinned by implicit physical knowledge and causal dynamics \cite{pols2024redesigning} that cannot be effectively captured by visual features alone. To facilitate research on such complex and cognitively rich scenarios, we introduce PhysLab, the first multi-granularity annotated video dataset specifically designed for physics experiment instruction.\par

As shown in Figure \ref{fig:intro}, PhysLab focuses on four representative undergraduate physics experiments, covering a diverse set of scientific instruments and typical student behaviors. The dataset contains 620 long-form videos (totalling 31 hours), each documenting authentic student experiment sessions. Videos are segmented into action clips, and richly annotated with hierarchical labels including action categories, object locations, interaction relationships, and operation sequences. These annotations provide a unified platform for multi-level or multi-task learning. In addition, the dataset incorporates structured metadata encoding experimental types, procedural steps, and tools used. This facilitates advanced tasks such as procedural modeling and experimental state reasoning. A key concept in this context is the process model \cite{peterson1977petri}, which refers to a structured representation of the experimental workflow, capturing the logical and temporal constraints among operations. It helps in understanding how experiments unfold over time, supporting reasoning about progress, correctness, and deviations during task execution.\par

To systematically assess the challenges introduced by PhysLab, we conducted extensive benchmark experiments spanning multiple mainstream visual parsing tasks. The evaluation results highlight the dataset’s complexity and expose substantial limitations in current methods, particularly in terms of insufficient visual context fusion, lack of prior knowledge, and pronounced inter-class performance disparity. We believe PhysLab offers a valuable foundation for advancing research in complex behavior modeling, improving educational process analysis, and bridging the gap between visual understanding and intelligent educational systems.\par

\section{Related work}


\subsection{Temporal Parsing Datasets}

Temporal parsing datasets focus on modeling the temporal dynamics of videos and support tasks such as action classification, temporal action localization, and action segmentation \cite{sun2022human, wang2023temporal}. Early datasets such as UCF101 \cite{soomro2012ucf101}, HMDB51 \cite{kuehne2011hmdb}, and Kinetics \cite{carreira2017quo} contain a variety of short and edited clips depicting everyday human actions, and have been widely adopted for action classification tasks. However, these datasets offer limited diversity in action types and lack well-defined temporal boundaries. To address this, subsequent datasets like THUMOS14 \cite{idrees2017thumos}, ActivityNet1.3 \cite{caba2015activitynet}, and FineAction \cite{liu2022fineaction} introduced untrimmed videos with denser temporal annotations, enabling research on action detection and segmentation.\par

More recently, as the community shifts towards modeling structured activities, several procedural video datasets have emerged. Datasets such as Breakfast \cite{kuehne2014language}, COIN \cite{tang2019coin}, and Assembly101 \cite{sener2022assembly101} focus on multi-step instructional tasks with well-defined goals, supporting the study of procedural understanding. In addition, datasets like EgoPER \cite{lee2024error}, CaptainCook4D \cite{peddi2024captaincook4d}, and IndustReal \cite{schoonbeek2024industreal} incorporate annotations of execution errors and deviations, facilitating research on anomaly detection in task workflows. 
Despite recent progress, existing datasets lack explicit procedural guidance for modeling complex workflows and are largely confined to narrow domains such as cooking and mechanical assembly \cite{tan2021comprehensive}. 
Datasets tailored to student laboratory experiments remain scarce, yet they are essential for advancing research at the intersection of visual perception and intelligent educational systems \cite{piloto2022intuitive}.\par

\vspace{-5pt}
\subsection{Spatial Parsing Datasets}

Spatial parsing datasets focus on object recognition and relational reasoning within static frames. General object detection datasets, such as COCO \cite{lin2014microsoft} and Objects365 \cite{shao2019objects365}, cover daily scenes. Domain-specific datasets, such as TJU-DHD \cite{pang2020tju} for autonomous driving, MVTec AD \cite{bergmann2019mvtec} and Read-IAD \cite{wang2024real} for industrial inspection, and HIOD \cite{hu2023object} for medical devices, provide high-quality samples that address challenges like occlusion, lighting variations, and scale diversity. Building upon object detection, instance segmentation datasets including ADE20K \cite{zhou2017scene} and COCONut \cite{deng2024coconut} offer pixel-level annotations, enabling finer-grained visual understanding.\par

Beyond object recognition tasks, datasets for HOI detection and scene graph generation aim to capture semantic relationships among objects, enabling more comprehensive contextual and relational reasoning \cite{han2024survey}. Representative datasets such as HICO-DET \cite{chao2018learning}, V-COCO \cite{gupta2015visual}, Visual Genome \cite{krishna2017visual}, and Open Images V4/V6 \cite{kuznetsova2020open} provide comprehensive annotations of both human-object and object-object interactions. Despite these advances, most existing datasets lack multi-granularity behavior annotations and task-level semantic alignment \cite{sener2022assembly101}, which limits their effectiveness in complex procedural modeling and causal behavior inference \cite{bi2021procedure}. In contrast, PhysLab integrates comprehensive temporal and spatial annotations, enabling diverse tasks ranging from video-level temporal action segmentation to frame-level object relational reasoning.\par

\vspace{-5pt}
\section{PhysLab dataset}

\subsection{Data Collection}
We selected four representative university-level physics experiments as the basis for data collection, covering the following experimental tasks: spectrometer-based angle measurement, moment of inertia determination for a rigid body, surface tension measurement, and object density measurement. To ensure the dataset’s authenticity and research value, the following considerations were incorporated during the data acquisition process:\par

\begin{itemize}
	\item \textbf{Task and behavioral diversity.} The chosen experiments span a variety of tasks, instruments, and manipulation patterns, covering diverse operational procedures and HOIs.
	
	\item \textbf{Environmental variability.} Video recordings were captured under varied conditions, including multiple camera viewpoints (three distinct angles), focal lengths, lighting setups, time-of-day variations, and laboratory environments. This diversity enhances the robustness and generalizability of the dataset across real-world deployment scenarios \cite{paulin2023review}.
	
	\item \textbf{Standardized experimental protocols.} All procedures strictly adhered to the official university physics experiment manuals, which provide detailed descriptions of objectives, apparatus specifications, procedural steps, data logging requirements, and safety guidelines. This ensures procedural consistency, reproducibility, and pedagogical validity.
	
	\item \textbf{Student autonomy and execution flexibility.} Each experiment was independently performed by students without external intervention, allowing for natural variations in execution order, action duration, and tool manipulation. This setting captures both successful operations and procedural errors, thereby supporting downstream tasks such as anomaly detection, procedural compliance analysis, and modeling of learning behaviors \cite{lee2024error}.
\end{itemize}

For video acquisition, we used the Newmine Q40 high-definition camera, which supports 4K resolution (3840×2160), a frame rate of 30 FPS, and a 90-degree distortion-free lens. This device delivers excellent image clarity, wide field-of-view coverage, and stable long-duration performance, making it well-suited for continuous laboratory deployment and autonomous data collection.\par

Following a rigorous filtering and cleaning process, we compiled a total of 620 experimental videos, with durations ranging from 1 to 8 minutes per video. In total, the dataset comprises approximately 31 hours of annotated video content, forming a high-quality and diverse corpus of procedural physics experiments. This resource is well-suited for visual parsing and experimental process modeling.\par

\subsection{Data Annotations}

\begin{figure*}
	\includegraphics[width=1.0\textwidth]{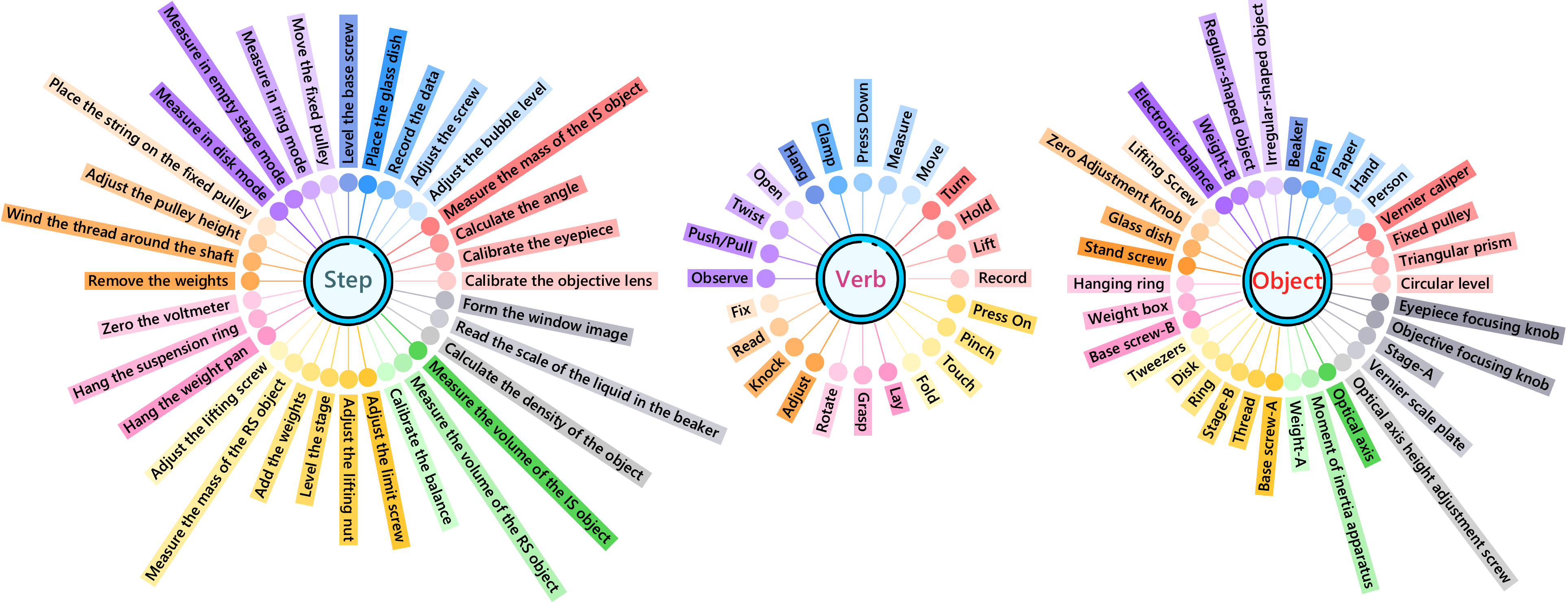}
	\caption{Taxonomy of experimental steps, interaction verbs, and objects in the PhysLab dataset.}
	\label{fig:stas}
\end{figure*}

In collaboration with the physics experiment center, we established a professional annotation team and adopted a ``multi-annotator \& multi-round verification'' protocol \cite{yang2024zhongjing} to ensure high annotation quality and consistency. Each sample was first annotated by an assigned annotator, then iteratively refined through reviews by multiple annotators. Both temporal and spatial annotations were performed to support multi-granularity visual parsing tasks. The annotated category schema is illustrated in Figure \ref{fig:stas}, with detailed quantitative statistics provided on our open-source website. \par

\textbf{Temporal Annotations.} For each video, we annotated the category, start time, and end time of each operational step along the timeline. These annotations facilitate tasks such as action classification in segmented videos and action localization in untrimmed videos \cite{ding2023temporal}. All labels strictly adhere to standardized procedures outlined in official experimental manuals, ensuring semantic consistency across the dataset. To enhance annotation objectivity, particularly in delineating precise temporal boundaries, each video was independently labeled by two experienced annotators using the ELAN \cite{wittenburg2006elan} tool. ELAN’s multi-tier timeline structure enables detailed, hierarchical descriptions of complex procedural behaviors. Following independent annotations, results were merged and manually reconciled to address discrepancies and ensure temporal coherence. In total, 3,873 action instances were annotated, each averaging approximately 20 seconds in duration and spanning 32 distinct types of experimental steps. These comprehensive annotations provide a robust foundation for procedural behavior analysis.\par

\textbf{Spatial Annotations.} To support spatial parsing, we extracted keyframes from the experimental videos and eliminated redundant frames using image similarity filtering techniques, resulting in a diverse subset of 4,500 high-quality frames. On these frames, we annotated bounding boxes for 34 categories of experimental instruments and objects, defined by the official equipment specifications in the experimental manuals. In addition, 24 types of interaction verbs were used to describe the functional relationships between the operator and the instruments, based on the semantics of experimental actions. Each annotated frame was thus structured into a triplet of the form <Operator, Interaction Verb, Instrument>. This annotation design enables HOI detection tasks \cite{yu2023hierarchical} and supports the construction of interaction graphs \cite{zhu2024calibration} for advanced reasoning and procedural understanding. Given the high density of objects, frequent interactions, and annotation complexity typical of physics experiment scenarios, a dedicated labeling team consisting of 5 postgraduate students and 18 undergraduate students carried out the annotation process using the Labelme tool. To ensure annotation quality, we implemented multi-round cross-checking and sample-based review procedures \cite{zhou2025browsecomp}.\par

\subsection{Data Statistics}

With the advancement of visual parsing technologies, research datasets have progressively evolved from capturing simple semantic scenes to encompassing more complex and cognitively demanding procedural task videos \cite{shen2024progress}. Table \ref{tab:statics} presents a summary and comparison of representative procedural video datasets, highlighting the advantages and distinctive features of the PhysLab dataset. \par

\begin{table*}
	\caption{Comparison of procedural visual parsing video datasets. AC: action classification, TAPG: temporal action proposal generation, AL: action alignment, AS: action segmentation, OD: object detection, IS: instance segmentation, OR: occlusion restoration, HOID: HOI detection, Flex.: more than one single execution order for the task, PEs: procedural errors.}
	\label{tab:statics}
	\centering
	\small
	\begin{tabular}{c|c|c|cccccccc|cc|cc}
		\toprule 
		& & & \multicolumn{8}{c|}{\textbf{Tasks}} & \multicolumn{2}{c|}{\textbf{Complexity}} & \multicolumn{2}{c}{\textbf{Dataset Size}} \\
		\textbf{Dataset} & \textbf{Year} & \textbf{Domain/Environment} & AC & TAPG & AL & AS & OD & IS & OR & HOID & Flex. & PEs & Videos & Hours\\
		\midrule
		Breakfast \cite{kuehne2014language} & 2014 & Cooking & \textcolor{green}{\cmark} & \textcolor{green}{\cmark} & \textcolor{green}{\cmark} & \textcolor{green}{\cmark}  & \textcolor{red}{\xmark} & \textcolor{red}{\xmark} & \textcolor{red}{\xmark} &\textcolor{red}{\xmark} & \textcolor{red}{\xmark} & \textcolor{red}{\xmark} & 1.7K & 77.0\\
		Epic-Kitchens \cite{damen2018scaling} & 2018 &  Cooking & \textcolor{green}{\cmark} & \textcolor{green}{\cmark} & \textcolor{green}{\cmark}  & \textcolor{green}{\cmark} & \textcolor{red}{\xmark} & \textcolor{red}{\xmark} & \textcolor{red}{\xmark} &\textcolor{red}{\xmark} & \textcolor{red}{\xmark} & \textcolor{red}{\xmark} & 432 & 55.0\\
		CorssTask \cite{zhukov2019cross} & 2019 & Daily Life & \textcolor{green}{\cmark} & \textcolor{green}{\cmark} & \textcolor{green}{\cmark} & \textcolor{green}{\textcolor{green}{\cmark}}  & \textcolor{red}{\xmark} & \textcolor{red}{\xmark} & \textcolor{red}{\xmark} &\textcolor{red}{\xmark} & \textcolor{red}{\xmark} & \textcolor{red}{\xmark}&  4.7K & 375.0\\
		COIN \cite{tang2019coin} & 2019 &  Daily Life & \textcolor{green}{\cmark} & \textcolor{green}{\cmark} & \textcolor{green}{\cmark} & \textcolor{green}{\cmark}  & \textcolor{red}{\xmark} & \textcolor{red}{\xmark} & \textcolor{red}{\xmark} &  \textcolor{red}{\xmark} & \textcolor{red}{\xmark} & \textcolor{red}{\xmark} & 11.8K & 476.0 \\
		Assembly101 \cite{sener2022assembly101} & 2022 & Toy Assembly & \textcolor{green}{\cmark} & \textcolor{green}{\cmark} & \textcolor{green}{\cmark} & \textcolor{green}{\cmark}  & \textcolor{red}{\xmark} & \textcolor{red}{\xmark} & \textcolor{red}{\xmark} &\textcolor{red}{\xmark} & \textcolor{green}{\cmark} & \textcolor{green}{\cmark} &  1.0K & 167.0\\
		HA4M \cite{cicirelli2022ha4m} & 2022 & Industrial Assembly & \textcolor{green}{\cmark} & \textcolor{green}{\cmark} & \textcolor{green}{\cmark} & \textcolor{green}{\cmark}  & \textcolor{red}{\xmark} & \textcolor{red}{\xmark} & \textcolor{red}{\xmark} &  \textcolor{red}{\xmark} & \textcolor{green}{\cmark} & \textcolor{red}{\xmark} & 217 & 5.9\\
		ATTACH \cite{aganian2023attach} & 2023 & Furniture Assembly & \textcolor{green}{\cmark} & \textcolor{green}{\cmark} & \textcolor{green}{\cmark} & \textcolor{green}{\cmark} & \textcolor{red}{\xmark} & \textcolor{red}{\xmark} & \textcolor{red}{\xmark} & \textcolor{red}{\xmark} & \textcolor{green}{\cmark} & \textcolor{red}{\xmark} &  378 & 17.2\\
		ATA \cite{ghoddoosian2023weakly} & 2023 & Toy Assembly & \textcolor{green}{\cmark} & \textcolor{green}{\cmark} & \textcolor{green}{\cmark} & \textcolor{green}{\cmark} & \textcolor{green}{\cmark} & \textcolor{red}{\xmark} & \textcolor{red}{\xmark} & \textcolor{red}{\xmark} & \textcolor{green}{\cmark} & \textcolor{red}{\xmark} & 1.2K & 24.8\\
		IndustReal \cite{schoonbeek2024industreal} & 2024 & Toy Assembly & \textcolor{green}{\cmark} & \textcolor{green}{\cmark} & \textcolor{green}{\cmark} & \textcolor{green}{\cmark} & \textcolor{green}{\cmark} & \textcolor{red}{\xmark} & \textcolor{red}{\xmark} & \textcolor{red}{\xmark} & \textcolor{green}{\cmark} & \textcolor{green}{\cmark} &  84 & 5.8\\
		EgoPER \cite{lee2024error} & 2024 & Cooking & \textcolor{green}{\cmark} & \textcolor{green}{\cmark} & \textcolor{green}{\cmark} &  \textcolor{green}{\cmark} & \textcolor{green}{\cmark} & \textcolor{red}{\xmark} & \textcolor{red}{\xmark} & \textcolor{red}{\xmark} & \textcolor{green}{\cmark} & \textcolor{red}{\xmark} &  386 & 28.0\\
		\midrule
		PhysLab (Ours) & 2025 & Physics Experiment & \textcolor{green}{\cmark} & \textcolor{green}{\cmark} & \textcolor{green}{\cmark} & \textcolor{green}{\cmark} & \textcolor{green}{\cmark} & \textcolor{green}{\cmark} & \textcolor{green}{\cmark} & \textcolor{green}{\cmark} & \textcolor{green}{\cmark} & \textcolor{green}{\cmark}& 620 & 31.0\\
		\bottomrule
	\end{tabular}
\end{table*}

Early datasets such as Breakfast \cite{kuehne2014language} and CrossTask \cite{zhukov2019cross} primarily focus on routine daily activities. While these datasets are relatively inexpensive to collect, the procedural content they offer is often simplistic, lacking the flexibility, variability, and error-prone behaviors that characterize real-world task execution. Consequently, their applicability for modeling complex task dynamics or analyzing execution uncertainties remains limited. \par

More recently, research has shifted toward industrial and simulated assembly scenarios, as seen in datasets such as Assembly101 \cite{sener2022assembly101}, HA4M \cite{cicirelli2022ha4m}, ATTACH \cite{aganian2023attach}, and IndustReal \cite{schoonbeek2024industreal}. These datasets feature more intricate task flows, typically involving furniture, toy, or industrial device assembly. They include annotations at both temporal and spatial levels, which significantly enhance their descriptive capacity and support comprehensive visual analysis. Building upon this trend, the PhysLab dataset we proposed is the first to target the domain of smart education. It comprises 620 real-world physics experiment videos performed by university students, totaling approximately 31 hours of footage, which is a considerable duration for datasets in this field. More importantly, PhysLab introduces the following distinctive features: \par

\begin{itemize}
	\item \textbf{High Task Authenticity.} Collected directly from real-world experimental environments, the dataset captures realistic student behaviors, including mistakes, incorrect operation sequences, and task deviations. This realism offers valuable insights into the natural variability of experimental learning processes.
	
	\item \textbf{High Process Flexibility.} Different students often adopt diverse operation orders and execution styles when performing the same experiment. Such variability facilitates the modeling of flexible and dynamic process structures, which is particularly beneficial for research in behavior understanding and process generalization.
	
	\item \textbf{Rich Annotation Granularity.} Beyond precise temporal boundaries for actions and object-level spatial labels, the dataset includes 24 categories of interaction verbs. These are used to construct structured HOI triplets that capture detailed interaction dynamics between students and instruments. These multi-granularity annotations significantly enhance the dataset’s value for multi-task modeling, such as action recognition, HOI detection, and process mining.
\end{itemize}

In summary, PhysLab stands out for its comprehensive coverage of experimental procedures, behavioral complexity, and multi-granularity annotations, making it a high-quality and multi-purpose dataset well suited for downstream applications in experimental process monitoring and visual process analysis.\par

\section{Selected Tasks}

As summarized in Table \ref{tab:statics}, PhysLab offers rich, diverse, and multi-granularity annotations, enabling a wide range of visual analysis tasks. In this section, we benchmark two representative settings with complementary emphases: video-level action recognition, which evaluates a model’s capacity to leverage temporal context, and image-level HOI detection, which assesses spatial reasoning in complex experimental environments. This dual-focus evaluation is intended to highlight the distinct challenges of visual parsing in physics experiment scenarios from both temporal and spatial perspectives. For additional tasks and benchmarks, please refer to the official open-source repository.\par

\vspace{-2mm}
\subsection{Action Recognition}

\textbf{Task Formulation.} The objective of action recognition is to locate and classify actions from untrimmed videos. In the context of procedural analysis \cite{li2019weakly, lu2021weakly}, this task is divided into two subtasks: 

\begin{itemize}
	\item \textbf{Action Alignment.} In this task, the transcript (i.e., the sequence of action labels) is known for a given test video, and the model is required to predict the temporal boundaries of each action segment to align with the ground truth (GT).
	
	\item \textbf{Action Segmentation.} In this more challenging task, no transcript is available. The model must simultaneously segment the temporal boundaries of actions and accurately predict the corresponding action labels.
	
\end{itemize}

\textbf{Evaluation Metrics.} We adopt two standard metrics: Mean over Frames (MoF) and Intersection over Union (IoU) \cite{zou2024weakly}. MoF measures the percentage of video frames for which the predicted action label matches the GT. IoU quantifies the average overlap between the predicted and GT boundary sets for each action. The two metrics are defined in Equation(\ref{eq-mof}) and Equation(\ref{eq-iou}), respectively. 

\begin{equation}
	\frac{1}{T} \sum_{i=1}^{T}(\hat{y_i}=y_i),
	\label{eq-mof}
\end{equation}

\begin{equation}
	\frac{1}{A} \sum_{a}\left| G_a \cap D_a \right|/\left| G_a \cup D_a\right|,
	\label{eq-iou}
\end{equation}

\noindent where $\hat{y}_i$ and $y_i$ denote the predicted and GT action labels of frame $i$, and $T$ is the total number of frames in the video. $G_a$ and $D_a$ represent the GT and predicted set of frames labeled with action $a$, and $A$ is the total number of distinct action categories.\par

\textbf{Results.} We evaluated four representative methods across three datasets: PhysLab, Breakfast, and CrossTask, with the latter two selected for being the authoritative and widely adopted procedural video benchmarks. As summarized in Table \ref{tab:ar}, these methods show overall inferior performance on PhysLab compared to the other two datasets, particularly in terms of MoF. This performance gap can be attributed to several factors inherent to PhysLab: the presence of diverse experimental actions, frequent occlusions, complex human-object interactions, and high visual similarity between distinct operations. These characteristics substantially increase the difficulty of both frame-level action classification and temporal boundary localization.\par

Moreover, PhysLab exhibits a stronger capacity for distinguishing model performance. For the action alignment task, the IoU gap between the best and worst-performing models on PhysLab reaches 21.2, whereas the corresponding gaps on Breakfast and CrossTask are only 5.0 and 3.1, respectively. A similar trend is observed in the action segmentation task, where PhysLab presents an IoU gap of 20.3, again notably larger than those on the other datasets. These results highlight that the complex visual and procedural characteristics of PhysLab not only introduce greater challenges for action recognition, but also provide a more rigorous benchmark for assessing model robustness and generalization capabilities.\par

\begin{table}[htb]
	\caption{Comparison of action alignment and action segmentation performance on PhysLab, Breakfast, and CrossTask. All reported values are multiplied by 100 for readability.}
	\label{tab:ar}
	\centering
	\setlength{\tabcolsep}{4.3pt}
	\small
	\begin{tabular}{cccccccc}
		\toprule
		& & \multicolumn{2}{c}{\textbf{PhysLab}} & \multicolumn{2}{c}{\textbf{Breakfast}} & \multicolumn{2}{c}{\textbf{CrossTask}}\\
		\cmidrule(lr){3-4} \cmidrule(lr){5-6} \cmidrule(lr){7-8}
		\textbf{Task} & \textbf{Method}  & MoF & IoU & MoF & IoU & MoF & IoU\\
		\midrule
		\multirow{4}{*}{Alignment} & CDFL \cite{li2019weakly}& 22.8 & 8.0 & 63.0 & 45.8 & 46.7 & 17.2\\
		& TASL \cite{lu2021weakly} & 30.5 & 20.4 & 65.8 & 49.9 &  57.1 & 19.1\\
		& POC \cite{lu2022set}& 36.0 & 28.0 & 56.1 & 46.7 & 53.3 & 18.9\\
		& AL-PKD \cite{zou2024weakly} & 41.5 & 29.2 & 67.6 & 50.8 & 62.7 & 20.3\\
		\midrule
		\multirow{4}{*}{Segmentation} & CDFL \cite{li2019weakly}& 12.0 & 5.7 & 50.2 & 33.7 & 32.5 & 11.8\\
		& TASL \cite{lu2021weakly} & 28.7 & 19.7 & 49.9 & 36.6 & 42.7 & 14.9\\
		& POC \cite{lu2022set} & 32.2 & 25.1 & 47.1 & 39.4 & 44.1 & 16.3\\
		& AL-PKD \cite{zou2024weakly} & 38.7 & 26.0 & 51.5 & 39.7 & 44.7 & 18.6\\
		\bottomrule
	\end{tabular}
\end{table}

\vspace{-2mm}
\subsection{Human-Object Interaction Detection}

\textbf{Task Formulation.} This task involves identifying interaction triplets from images, each consisting of the bounding box of the human subject, the category and bounding box of the interacting object, and the interaction verb describing their relationship \cite{han2024survey}.\par

\textbf{Evaluation Metrics.} We adopt the standard mean Average Precision (mAP) under the Full, Rare, and Non-Rare settings, following the HICO-DET evaluation protocols \cite{chao2018learning}. A predicted interaction triplet is considered a true positive (TP) if and only if all of the following conditions are met:

(1) The predicted category labels for both the human and object bounding boxes are correct.\par
(2) The predicted human and object boxes each have an IoU (spatial-level) with the corresponding GT box greater than 0.5.\par
(3) The predicted interaction verb matches the GT label.\par

Given these criteria, the Average Precision (AP) is computed for each interaction category, and overall mAP is defined as:\par

\begin{equation}
	mAP = \frac{1}{C} \sum_{i=1}^{C}AP_i,
	\label{eq-map}
\end{equation}

\noindent where $C$ is the number of interaction categories, and $AP_i$ denotes the average precision for category $i$.\par

\textbf{Results.} We evaluated 10 representative HOI detection methods on PhysLab and HICO-DET, adopting the latter as the reference benchmark since other datasets listed in Table \ref{tab:statics} lack HOI annotations. Results are summarized in Table \ref{tab:hoi}. Following \cite{liao2022gen, chen2025focusing}, the backbones are organized into three groups: ResNet-50, ResNet-101, and Swin-L. Overall, model performance on PhysLab is generally higher than on HICO-DET, primarily because PhysLab focuses on structured and domain-specific interactions in physics experiments, whereas HICO-DET covers diverse and open-ended scenarios. Despite this performance advantage, PhysLab introduces two unique challenges. First, its pronounced category imbalance results in significant performance shifts across classes. Unlike HICO-DET, where most models consistently perform better on Non-rare categories, models on PhysLab exhibit divergent trends. For instance, STIP \cite{zhang2022exploring} and LOGICHOI \cite{li2023neural} perform better on Non-rare interactions, while OCN \cite{yuan2022detecting} achieves superior performance on Rare categories (achieving  68.01 mAP on Rare classes vs. 51.10 on Non-rare). Notably, GEN-VLKT$_s$ \cite{liao2022gen} exhibits a 17.42 mAP gap between the two sets, highlighting the increased generalization demands posed by PhysLab under imbalanced label distributions.\par

\begin{table}[tb]
	\caption{Comparison of human-object interaction detection performance on PhysLab and HICO-DET. All reported values are multiplied by 100 for readability.}
	\label{tab:hoi}
	\centering
	\setlength{\tabcolsep}{2.8pt}
	\small
	\begin{tabular}{ccccccc}
		\toprule
		& \multicolumn{3}{c}{\textbf{PhysLab}} & \multicolumn{3}{c}{\textbf{HICO-DET}}\\
		\cmidrule(lr){2-4} \cmidrule(lr){5-7}
		\textbf{Method} & Full & Rare & Non-Rare &  Full & Rare & Non-Rare\\
		\midrule
		STIP \cite{zhang2022exploring}  & 62.62 & 61.00 & 62.73 & 32.22  & 28.15 & 33.43 \\
		GEN-VLKT$_s$ \cite{liao2022gen}  & 58.71 & 75.00 & 57.58 & 33.75 & 29.25 & 35.10\\
		OCN \cite{yuan2022detecting}  & 52.19 & 68.01 & 51.10 & 30.91 & 25.26 & 32.51 \\
		LOGICHOI \cite{li2023neural}  & 53.34 & 50.97 & 62.50 & 35.47 & 32.03 & 36.22\\
		TED-Net \cite{wang2024ted}  & 60.97 & 69.89 & 60.35 & 34.00 & 29.88 & 35.24\\
		SOV-STG-S \cite{chen2025focusing}  & 45.42 & 34.17 & 49.50 & 33.80 & 29.28 & 35.15\\ 
		\midrule 
		OCN \cite{yuan2022detecting}  & 49.50 & 50.00 & 49.46 & 31.43& 25.80& 33.11\\
		UPT \cite{zhang2022efficient}  & 65.28 & 63.47 & 65.43 & 32.31 & 28.55 & 33.44\\
		GEN-VLKT$_l$ \cite{liao2022gen}  & 60.19 & 64.25 & 69.91 & 34.95 & 31.18 & 36.08 \\
		SOV-STG-L \cite{chen2025focusing} &  52.35 & 46.59 & 58.44 & 35.01 & 30.63 & 36.32\\
		\midrule
		PViC \cite{zhang2023exploring}  & 71.42 & 68.47 & 71.62 & 44.32 & 44.61 & 44.24\\
		FGAHOI \cite{ma2024fgahoi}  & 60.79 & 54.89 & 67.23 & 37.18 & 20.71 & 39.11\\
		SOV-STG-Swin-L \cite{chen2025focusing} & 56.48 & 45.23 & 58.44 & 43.35 & 42.25 & 43.69\\
		\bottomrule
	\end{tabular}
\end{table}

Second, model performance exhibits greater relative volatility across models in each evaluation subset. Specifically, within the Full, Rare, and Non-Rare subsets of PhysLab, the performance gap between the best and worst models (calculated as the percentage difference normalized by the worst model’s score) reaches 57.2\%, 119.5\%, and 44.8\%, respectively. These gaps are significantly higher than those observed in HICO-DET, which are 43.4\%, 115.4\%, and 20.3\%, respectively. This suggests that PhysLab’s interaction scenarios not only amplify the strengths and weaknesses of different model architectures, but also better reveal their varying adaptability to distinct interaction types and category frequencies \cite{chao2018learning}. As described in \cite{liu2019large, chan2024auxiliary}, the greater inter-class performance differences in PhysLab make it a more challenging benchmark \cite{chan2024auxiliary}. These characteristics underscore PhysLab’s value in promoting the development of more robust and advanced HOI detection algorithms \cite{zou2021end}, particularly in experimental or instructional contexts \cite{hu2024noisy}.\par

\section {Conclusion and Future Work}

In this paper, we introduce PhysLab, a richly annotated dataset designed to advance multi-granularity visual parsing of physical experiment processes. PhysLab fills a critical void at the intersection of computer vision, multimedia, and science education by embedding visual recognition tasks within authentic experimental contexts.\par

We will actively expand the dataset by annotating data for six additional representative physics experiments. Concurrently, we plan to collect experimental procedures in chemistry and biology, thereby increasing domain diversity, enabling cross-domain adaptation tests, and supporting cross-disciplinary generalization. To enable more robust and context-aware modeling of experimental processes, we aim to  integrate multimodal signals such as audio, textual instructions, and sensor data, and explore cross-modal alignment and fusion strategies to better capture the semantic structure of experimental tasks. All versions of the dataset, along with benchmark protocols and evaluation results, will be continuously maintained and released through our open-source platform.\par

\begin{acks}
This work was supported in part by the National Major Science and Technology Projects of China award [2022ZD0119501], National Natural Science Foundation of China award [52374221], and Science and Technology Development Fund of Shandong Province of China award [ZR2023MF097].
\end{acks}

\bibliographystyle{ACM-Reference-Format}
\balance
\bibliography{sample-base}

\end{document}